\documentclass[a4paper,fleqn,]{cas-sc}

\usepackage{caption}
\usepackage{float}
\usepackage[numbers]{natbib}
\usepackage{graphicx}
\usepackage{orcidlink}
\usepackage{multicol} 
\usepackage{booktabs}
\usepackage{tabularx}
\usepackage{threeparttable}
\usepackage{array}
\usepackage{amssymb}
\def\tsc#1{\csdef{#1}{\textsc{\lowercase{#1}}\xspace}}
\tsc{WGM}
\tsc{QE}

\newcommand{\oid}[1]{\texorpdfstring{\,\orcidlink{#1}}{}}
\begin{document}
\let\WriteBookmarks\relax
\def\floatpagepagefraction{1}
\def\textpagefraction{.001}

\shorttitle{}    


\title [mode = title]{Hardware-Free Robotics Laboratories in Mixed Reality}  



%






\author[1]{Santiago Berrezueta-Guzman\oid{0000-0001-5559-2056}}[orcid=0000-0001-5559-2056]\cormark[1]
\ead{s.berrezueta@tum.de}
\author[1]{Habiba-Loai Khalil\oid{0009-0007-3085-5105}}[orcid=0009-0007-3085-5105]
\author[1]{Andrei Koshelev\oid{0009-0000-3786-4994}}[orcid=0009-0000-3786-4994]
\author[1]{Vanesa Metaj\oid{0009-0008-6576-346X}}[orcid=0009-0008-6576-346X]
\author[1]{Stefan Wagner\oid{0000-0002-5256-8429}}[orcid=0000-0002-5256-8429]




\affiliation[1]{organization={Technical University of Munich},
            addressline={}, 
            city={Heilbronn},
            country={Germany}}

\cortext[1]{Corresponding author}



\begin{abstract}
Teaching robotics relies on screen-based simulation, showing robot motion in an abstract coordinate frame rather than at real scale in the learner's own space, while access to physical hardware is limited by cost, safety, and scheduling constraints. 
We present \textit{MR-Robotics~LAB}, a mixed-reality (MR) platform that replays MATLAB-generated robot trajectories at real scale within the learner's physical environment. 
A browser-based service validates a MATLAB workspace file (\texttt{.mat}), normalizes units, and publishes a versioned JSON trajectory; a Unity application on a Meta Quest~3 then reproduces the authored joint configurations under position control and replays them at the declared frame rate within a physics-enabled scene that supports collision detection and end-effector grasping. 
A formative single-group evaluation with engineering students found that participants reported low setup effort ($M = 4.67$ on a 5-point scale) and perceived support for workspace understanding from multi-viewpoint inspection ($M = 4.56$), and 83\% of participants affirmed their willingness to use the platform in an introductory robotics course. 
The evaluation instrument records only perceived outcomes, without counterbalancing or a learning measure, so no comparative advantage over desktop simulation is claimed. 
The contribution is a reusable simulation-to-MR trajectory pathway and design guidance for hardware-free robot visualization in engineering education.
\end{abstract}




\begin{keywords}
Mixed reality \sep Robotics education \sep Robot trajectory visualization \sep 
MATLAB Robotics Toolbox \sep Immersive learning \sep Spatial reasoning \sep Engineering education \sep design-based research \sep extended reality
\end{keywords}

\maketitle
\begin{multicols}{2}

\section{Introduction}\label{I}

Nowadays, teaching and learning robotics involves simulation and visualization environments for trajectory development, kinematic analysis, and motion inspection.
Students design and validate robot programs in these environments, observing the output as 2D plots or flat-screen renderings. 
While these tools are effective for programming and kinematic validation, robotics education faces three compounding constraints: the spatial limitations of screen-based simulation tools, the cost of hardware, and the scarcity of physical resources for their allocation, which includes laboratory infrastructure, safety requirements, and the need to share limited robot time among students \cite{wiedmeyer2019robotics, rukangu2025virtual}.

As a result, the experiential component of robotics education, understanding robot behavior in a real, three-dimensional, physically meaningful context, remains limited.
This matters because spatial reasoning skills are strongly predictive of achievement in STEM disciplines and are malleable through spatially enriched training \cite{uttal2013malleability}, a development that screen-based simulation constrains. 

Mixed Reality (MR) offers a distinct alternative when physical robot access is limited, situating a virtual robot at real scale in the learner's physical environment and providing spatial correspondence that conventional screen-based simulation does not offer \cite{speicher2019mixed}. 
The learner can observe the robot's planned and programmed trajectory at real scale, in their actual surroundings, and reason about spatial relationships as they would in a physical setting.

In this paper, we present \textit{MR-Robotics LAB}\footnote{MR-Robotics LAB: \url{https://mr-lab.se.cit.tum.de/}}, a system that provides an MR simulation platform that enables students to visualize and interact with their robot trajectory programs in an immersive, physics-enabled environment. 
Students can upload their MATLAB workspace file (\texttt{.mat}) with the joint trajectory to the MR-Robotics LAB web platform, then place a virtual robot model in their physical surroundings using an MR headset (Meta Quest 3) (e.g., classroom, laboratory, or any available space), and execute the simulation. 

The platform replays the uploaded joint trajectory at room scale, preserving the uploaded joint configurations and replaying them at the declared frame rate encoded in the converted trajectory record. It operates within a physics-enabled mixed-reality scene, detecting collisions with both detected physical surroundings and spawned virtual objects. An interchangeable end-effector system enables physics-based grasping and manipulation of virtual objects during pick-and-place trajectories. The experience requires minimal configuration; the gap between completing a MATLAB script and observing its execution in MR requires only a few steps.

\subsection{Research Questions}

Based on the development, deployment, and formative evaluation of the platform, we address the following research questions (RQs). 

\textbf{RQ1.} How do students rate the effort and confidence required to move from an existing MATLAB trajectory file to room-scale MR playback, and what usability obstacles do they report?

\textbf{RQ2.} Which spatial properties of robot motion do students report becoming accessible to them under scale-correct MR playback, and how do they characterize these relative to their prior experience of screen-based visualization?

\textbf{RQ3.} Do students express willingness to adopt the platform within an introductory robotics course, and what extensions do they request?

\subsection{Contributions}

This work makes four contributions to the existing literature. 

1. We provide a browser-based pipeline that converts MATLAB trajectory files into a structured format for MR playback, allowing conversion without requiring a MATLAB installation at conversion time and providing secure, user-scoped storage.

2. We provide a physics-aware MR playback system that faithfully reproduces MATLAB-computed trajectories at real scale in the learner's physical environment, with real-time collision detection and physics-based object manipulation.

3. We designed a multi-robot MR simulation environment with a minimal-configuration interaction model, allowing multiple robot instances to be placed and observed simultaneously alongside on-demand joint telemetry, and to be grouped by direct spatial selection into user-defined systems whose members execute in parallel or in sequence under a single combined timeline.

4. We articulate and document the control-mode separation required when prescribed kinematic playback must coexist with a physically reactive scene, including the grasp-as-constraint decision and its consequences for what students observe. 

\section{Related Work}\label{RW}

\subsection{AR/VR/MR in Engineering Education}

The application of MR to engineering education has attracted substantial empirical attention over the past decade \cite{bacca2014augmented}. 
Systematic reviews of augmented reality (AR) in education indicate that immersive learning environments consistently improve student engagement and support the visualization of abstract or spatially complex concepts \cite{akccayir2017advantages, bacca2014augmented}. 
A study~\cite{akccayir2017advantages} identified enhanced interactivity and the ability to render otherwise unobservable phenomena as the primary pedagogical advantages of MR, while also noting that usability, specifically the complexity of the interaction model, constitutes the most frequently reported challenge.

In engineering education specifically, MR has been applied across disciplines, including electronics, structural analysis, and manufacturing. 
A systematic review about it found predominantly positive effects on student learning and engagement, while identifying technology acceptance and interaction complexity as recurring practical barriers \cite{alvarez2022augmented}. 

Additionally, spatially enriched tools in engineering contexts have been linked to improved conceptual understanding of three-dimensional structures and dynamic processes \cite{fogarty2018improving}. 
In robotics education, the study \cite{verner2015reorganizing} demonstrated that robotics laboratory activities can serve as a vehicle for spatial skill development in novice engineering students, providing direct motivation for situated robot simulation as a spatial learning tool.

However, a study found that, without structured scaffolding, immersive environments can introduce extraneous cognitive load, impairing learning outcomes relative to conventional slide-based instruction, underscoring the importance of deliberate pedagogical design in MR systems rather than immersion alone \cite{parong2018learning}.

\subsection{Robot Simulation Tools}

Robotics education uses simulation environments for trajectory development, kinematic analysis, and motion inspection. 
A systematic review of realistic simulators in educational robotics identified physics engine integration and 3D visual fidelity as the primary criteria for educational suitability \cite{camargo2021systematic}.

Established platforms, such as MATLAB Robotics Toolbox \cite{corke2017robotics}, ROS with RViz \cite{quigley2009ros}, Webots \cite{michel2004cyberbotics}, and CoppeliaSim (formerly V-REP \cite{rohmer2013v}) render three-dimensional robot models but present them on-screen rather than registered to the learner's physical workspace. 
When students learn about robots through screens, they risk failing to adequately understand the physical scale and real-world spatial relationships \cite{rukangu2025virtual}.

Prior work has applied AR specifically to robot programming. 
The study~\cite{quintero2018robot} demonstrated AR-enabled trajectory specification, virtual previews of robot motion, and on-screen reprogramming using a HoloLens with a 7-DOF manipulator, supporting visualization of robot parameters in the operator's physical space. 
Another study~\cite{dogangun2025rampa} extended this to programming-by-demonstration using a Meta Quest 3, enabling trajectory recording, visualization, and fine-tuning within the user's physical environment. 
Both systems target industrial programming workflows with physical robot hardware present; neither is designed for hardware-free educational use or for replaying student-authored simulation scripts.

Game engines, notably Unity and NVIDIA Isaac Sim, have also been applied to robot simulation, primarily in reinforcement learning and sim-to-real transfer \cite{tobin2017domain}, rather than educational visualization. Sibilska-Mroziewicz et al. \cite{sibilska2023analysis} compared VR visualization of robot kinematics directly against 2D plots and 3D screen animations in an engineering context, concluding that the immersive approach better supports analysis and design by allowing the viewer to inspect motion from multiple perspectives, which is a finding that motivates the MR approach of the present work.

Table~\ref{tab:comparison} compares representative desktop simulators and in-situ AR/MR robot interfaces along the dimensions most relevant to the present work. 
Desktop simulators such as MATLAB Robotics Toolbox, ROS with RViz, Webots, and CoppeliaSim provide three-dimensional robot visualization, while Webots and CoppeliaSim additionally support physics-enabled simulation. In contrast, \cite{quintero2018robot} and \cite{dogangun2025rampa} use AR to support robot programming in the user's physical workspace, but both are designed around interaction with physical industrial robots. 
MR-Robotics LAB combines hardware-free trajectory playback, room-scale MR visualization, and physics-enabled object interaction in an instructional workflow based on student-authored MATLAB trajectories.

\end{multicols}
\captionof{table}{Comparison of various robotics visualization and simulation systems.}
\label{tab:comparison}
\begin{tabular}{|p{4.5cm}|c|c|c|c|c|}

\hline
\textbf{System} &
\shortstack{\textbf{Trajectory}\\\textbf{vis./playback}} &
\shortstack{\textbf{Interactive}\\\textbf{physics}} &
\shortstack{\textbf{In-situ}\\\textbf{AR/MR}} &
\shortstack{\textbf{Room-}\\\textbf{scale}} &
\shortstack{\textbf{Physical robot}\\\textbf{required}} \\
\hline

MATLAB Robotics Toolbox \cite{corke2017robotics} &
\checkmark & -- & -- & -- & -- \\
\hline
ROS + RViz \cite{quigley2009ros} &
\checkmark & -- & -- & -- & -- \\
\hline
Webots \cite{michel2004cyberbotics} &
\checkmark & \checkmark & -- & -- & -- \\
\hline
CoppeliaSim \cite{rohmer2013v} &
\checkmark & \checkmark & -- & -- & -- \\
\hline
Quintero et al. \cite{quintero2018robot} &
\checkmark & -- & \checkmark & \checkmark & \checkmark \\
\hline
RAMPA \cite{dogangun2025rampa} &
\checkmark & -- & \checkmark & \checkmark & \checkmark \\
\hline
\rowcolor{gray!20}
\textbf{MR-Robotics LAB} &
\textbf{\checkmark} & \textbf{\checkmark} & \textbf{\checkmark} &
\textbf{\checkmark} & \textbf{--} \\
\hline
\end{tabular}
\vspace{2pt}
\parbox{0.95\textwidth}{\footnotesize
\textit{Note:} ``--'' indicates that the feature is not a primary function of the cited system.}       

\begin{multicols}{2}

\subsection{Embodied cognition and spatial learning}

Embodied cognition holds that reasoning about physical systems is grounded in sensorimotor interaction with the environment \cite{wilson2002six}, implying that spatial reasoning depends on how far a learning medium affords such engagement.
A study extended this to immersive technologies, arguing that engaging with a representation at real scale, by observing, walking around, and gesturing within a three-dimensional scene, recruits embodied resources that a perspective-projected rendering on a flat display does not elicit \cite{johnson2018immersive}.
Spatial ability is among the strongest predictors of STEM achievement, and training studies show these skills are malleable \cite{uttal2013malleability, wai2009spatial}.

Robotics laboratory activities have been proposed as a vehicle for spatial skill development precisely because they require students to reason about three-dimensional workspace geometry in a physically grounded context \cite{verner2015reorganizing}.
Recent evidence suggests that situated visualization may compensate for lower baseline spatial ability \cite{leins2026investigating}, underscoring the importance of evaluating spatial outcomes not only on average but across individual differences.

This theoretical framing motivates three specific design decisions in MR-Robotics LAB: (1) rendering the robot in the student's actual physical environment rather than in a neutral virtual room; (2) including real-time joint telemetry as a scaffold that connects abstract trajectory values to visible motion; and (3) minimizing interaction complexity to reduce extraneous cognitive load.

\section{The MR-Robotics LAB platform }\label{ARLab}

The MR-Robotics LAB operationalizes the affordances motivated in the \textit{related work} section through two coordinated subsystems: a web-based conversion service that transforms a MATLAB workspace (\texttt{.mat}) into a self-describing trajectory contract, and an MR runtime that reproduces that trajectory at real scale in the student's physical environment. 

\subsection{System Overview}

Figure~\ref{fig:system-overview} summarizes the end-to-end workflow of the MR-Robotics LAB, which comprises a web-based trajectory conversion service and an MR application developed in Unity and deployed on the Meta Quest 3. 

Students first export trajectory data from MATLAB as a workspace \texttt{.mat} file and upload it through the web platform, which validates the motion data, associates it with a selected robot model (and gripper), and publishes a standardized JSON trajectory to our Firebase cloud storage. 
The MR application in the headset periodically synchronizes with Firebase Storage to retrieve published trajectories. 
The trajectories are filtered by the robot model identifier selected within the headset, preventing kinematic mismatches between uploaded trajectory data and the instantiated robot. 
Finally, the student places their robot in the scene, selects their trajectory, and initiates playback to analyze its program and final output with the Virtual Robot.

The rendering pipeline uses Unity's Universal Render Pipeline (URP)~\cite{unity_urp}. 
The render order follows the standard URP sequence: \textit{opaque geometry} (robot body, physics objects) → \textit{transparent geometry} (holographic end-effector indicator) → \textit{composited onto the Meta Quest 3 passthrough camera feed}, so that virtual objects are correctly depth-composited with the real environment and the robot appears to stand on the actual floor surface. 

\end{multicols}

\begin{center}
  \includegraphics[width=\linewidth]{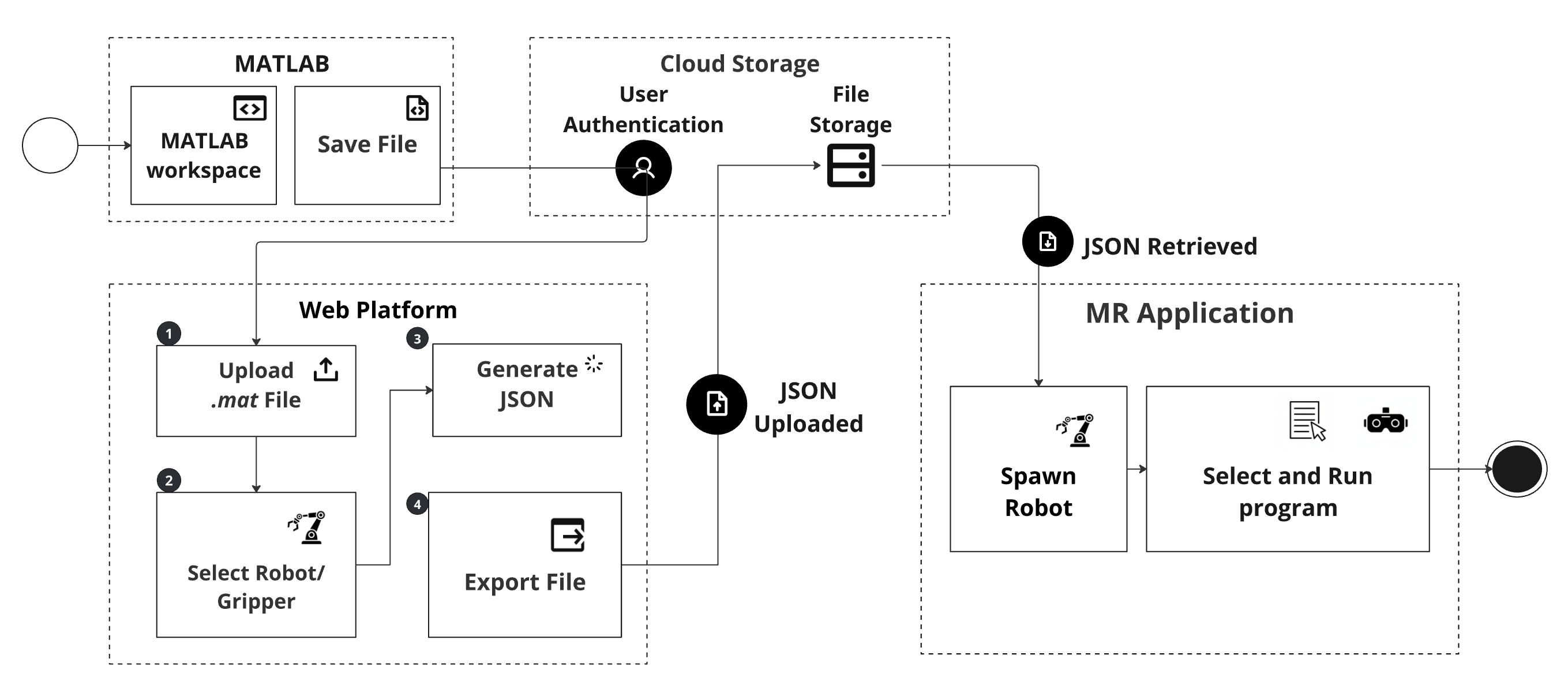}
  \captionof{figure}{High-level system overview of the MR-Robotics LAB workflow.}
  \label{fig:system-overview}
\end{center}

\begin{multicols}{2}

The virtual scene is rendered entirely within the student's physical environment using the Meta Quest 3's color passthrough feed. No virtual room or synthetic background is introduced; the robot appears to occupy the student's actual space, a classroom, laboratory, or any available area, at real scale, depth-composited with the passthrough image so that the robot appears to stand on the actual surface as shown in Figure~\ref{fig:robots-on-table}.

Table \ref{tab:robots} presents the available robots in our platform. Robot models are defined using the industry-standard Unified Robot Description Format (URDF)~\cite{quigley2009ros}, ensuring geometric and kinematic fidelity to real robot specifications \cite{kuka2025robots,universalrobots2025cobots}. All models are six-DOF serial manipulators. 
The UR5e was included to verify that the URDF-based model pipeline generalizes beyond a single manufacturer.

\captionof{table}{Robot models currently supported.}
\label{tab:robots}
\centering
\footnotesize
\setlength{\tabcolsep}{4pt}
\begin{tabular}{@{}lccc@{}}
\toprule
\textbf{Model} & \textbf{Class} &
\shortstack{\textbf{Payload}\textbf{(kg)}} &
\shortstack{\textbf{Reach}\textbf{(mm)}} \\
\midrule
\multicolumn{4}{@{}l}{\textit{KUKA}} \\
\quad KR3 R540         & Industrial   & 3   & 541  \\
\quad KR6 R900 sixx    & Industrial   & 6   & 901  \\
\quad KR16 R1610-2     & Industrial   & 16  & 1610 \\
\quad KR20 R1810-2     & Industrial   & 20  & 1813 \\
\quad KR120 R2500 Pro  & Industrial   & 120 & 2496 \\
\quad KR150 R2700-2    & Industrial   & 150 & 2701 \\
\quad KR10 R900-2      & Industrial   & 11  & 901  \\
\quad KR50 R2100       & Industrial   & 50  & 2101 \\
\quad KR70 R2100       & Industrial   & 70  & 2101 \\
\quad KR300 R2700-2    & Industrial   & 300 & 2701 \\
\quad KR560 R3100-2    & Industrial   & 560 & 3100 \\
\quad LBR iisy 15 R930 & Collaborative arm  & 15  & 930  \\
\addlinespace[2pt]
\multicolumn{4}{@{}l}{\textit{Universal Robots}} \\
\quad UR5e             & Collaborative arm  & 5   & 850  \\
\bottomrule
\end{tabular}
\vspace{2pt}

\end{multicols}

\begin{center}
  \includegraphics[width=0.65\linewidth]{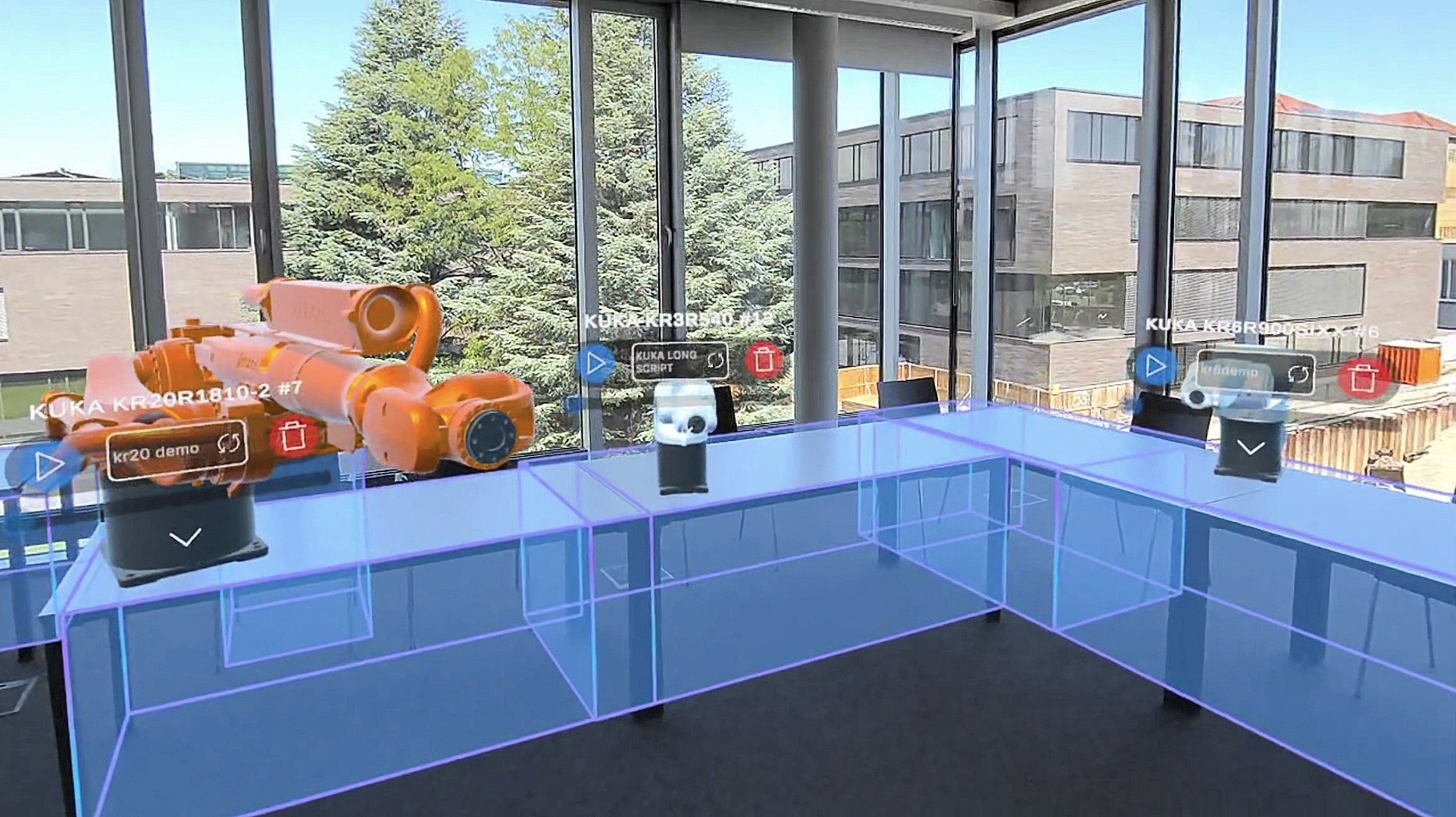}
  \captionof{figure}{Robots standing on the table in a classroom.}
  \label{fig:robots-on-table}
\end{center}

\begin{multicols}{2}

\subsection{Core Technical Challenges} 

To reach the end-to-end workflow explained above, we established eight principal challenges to structure the design and implementation of the system:  

\textit{1. Trajectory detection and validation:} MATLAB workspace files contain multiple numeric arrays with inconsistent variable names. 
The web platform must automatically identify the correct two-dimensional trajectory matrix, reject invalid shapes and non-finite values, and support manual selection when several workspace candidates are plausible.

\textit{2. Unit normalization and interchange contract:} Joint angles authored in radians must be converted to degrees for MR playback, and an optional seventh column must be interpreted as a gripper signal and mapped to binary open/closed states. 
The converter must emit a stable, versioned JSON representation with explicit metadata so the MR application can consume trajectories without manual reformatting.

\textit{3. Kinematic enrichment for visualization:} The interchange format must include per-frame joint velocities and display-oriented nominal torque proxies derived from the uploaded trajectory. These quantities must be estimated consistently from discrete samples and presented as qualitative teaching telemetry, not as measured or robot-specific dynamics for analysis or control.

\textit{4. Web deployment and cloud distribution:} The upload path must parse \texttt{.mat} files in the browser without requiring a MATLAB installation at conversion time, report actionable diagnostics when parsing or detection fails, and publish converted trajectories to cloud storage under the selected robot model so the headset can retrieve the correct script.

\textit{5. Trajectory fidelity:} The MR simulation must faithfully reproduce the robot's planned kinematic motion. Joint configurations must correspond to the MATLAB-computed trajectory, while playback timing follows the frame rate declared during conversion, so that the student observes the intended motion consistently in MR.

\textit{6. Physical realism:} The virtual robot must behave as a physical machine: detecting and responding to collisions with objects in its environment, stopping execution upon contact (like an emergency stop on a real system), and interacting with manipulable objects through its end-effector.

\textit{7. User experience:} The interaction model must be accessible to users without prior MR experience. Placement, control, and observation of the robot should require minimal training and no manual parameter configuration.

\textit{8. Minimal setup and accessibility:} The end-to-end workflow from trajectory authorship to MR visualization should be as short as possible, enabling the platform to function as a natural extension of the existing MATLAB-based curriculum rather than a separate tool requiring significant onboarding.

Challenges 1-4 are addressed by the web-based trajectory conversion pipeline, whereas challenges 5-8 are addressed by the MR application.

\subsection{Web-Based Trajectory Conversion}

The workflow begins in MATLAB, where students compute joint-space motion and save it as a workspace \texttt{.mat} file. The web conversion service takes that file and turns it into a trajectory that the MR runtime can play back without further interpretation. The problem it solves is one of mismatch: a MATLAB workspace is unstructured, and its conventions are implicit, whereas the runtime needs a known layout, declared units, and an explicit robot identifier. The service resolves this by validating the data, normalizing its conventions, and attaching metadata before publication, so that playback can treat the published file as authoritative. Figure~\ref{fig:web-overview} shows the full pipeline.

\end{multicols}

\begin{center}
  \includegraphics[width=\textwidth]{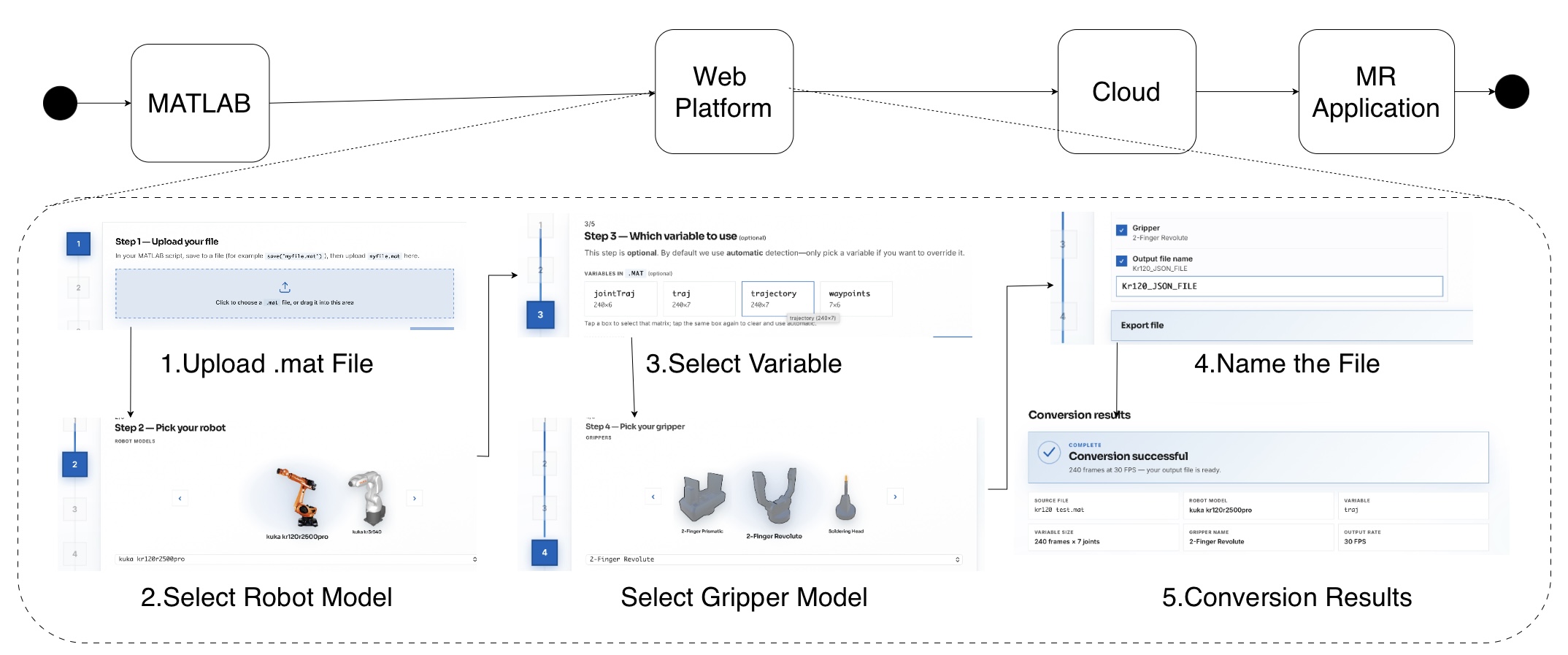}
  \captionof{figure}{Web-based trajectory conversion workflow.}
  \label{fig:web-overview}
\end{center}

\begin{multicols}{2}

A trajectory is a real matrix $\mathbf{Q} \in \mathbb{R}^{N \times d}$ with $N \ge 2$ temporally ordered samples and $d \in \{6,7\}$ columns. The samples contain no explicit timestamps; uniform timestamps are assigned during conversion from the declared frame rate. The first six columns are the arm joints in radians, in a fixed order that is the same in every frame; an optional seventh column carries the gripper signal. Because a workspace file may contain several variables, the service automatically locates the trajectory matrix from its dimensions and name and asks the student to choose when more than one candidate fits. It then checks that the matrix has a valid shape and contains only finite values, leaving the joint ordering untouched.

The validated trajectory is tagged with the selected robot model and,
where applicable, its gripper configuration. This metadata travels with
the trajectory and lets the MR application show each student only the
scripts that match the robot currently placed in the scene.

\subsection{Canonical Joint and Gripper Representation}

The selected matrix is transformed into a frame-based JSON trajectory. Arm joint angles are converted from radians to degrees according to $\Theta_{i,j}=\frac{180}{\pi}Q_{i,j}$ for $i=0,\ldots,N-1$ and $j=1,\ldots,6$, while preserving the column order established in MATLAB.

For an optional gripper signal $g_i = Q_{i,7}$, the converter first computes a threshold
\begin{equation}
\tau_g=\frac{\min_i g_i+\max_i g_i}{2},
\label{eq:gripper-threshold}
\end{equation}

which is then used to assign binary gripper states,
\begin{equation}
s_i=
\begin{cases}
1, & g_i>\tau_g \quad \text{(closed)},\\
0, & g_i\le\tau_g \quad \text{(open)}.
\end{cases}
\label{eq:gripper-bin}
\end{equation}

If the gripper signal is constant, the converter assigns $s_i=0$ for all frames. Gripper states are not interpolated during playback; a state change occurs only at a frame boundary where the binarized gripper signal changes.

Timing information is assigned from a declared frame rate $f$, yielding uniformly spaced frame timestamps $t_i=i\Delta t$, where $\Delta t$ is determined by the specified frame rate. Each published frame therefore contains the robot joint configuration, an optional gripper state, and an explicit timestamp, allowing the MR application to reconstruct the trajectory without inferring timing or units from the original MATLAB workspace.

\subsection{Derived Kinematic Channels and Cloud Publication}

Although the playback pose is driven by the uploaded joint positions, the published artifact also includes derived kinematic channels intended for student-facing telemetry. Let $\mathbf{q}_i \in \mathbb{R}^6$ denote the arm joint vector in radians at frame $i$, and $t_i$ the associated timestamp. Joint velocities are estimated by numerical differentiation of $\mathbf{q}_i$ with respect to time (central differences at interior frames and one-sided differences at the endpoints), then lightly smoothed to suppress high-frequency artefacts from discrete sampling. Joint accelerations are obtained by differentiating the smoothed velocity sequence and clipping extreme values to a bounded range so that occasional numerical spikes do not dominate the displayed signal.

A simplified per-joint torque proxy is then computed in the spirit of a diagonal dynamics model,
\begin{equation}
\begin{aligned}
  \hat{\tau}_{i,j} &=
  I_j\,\ddot{q}_{i,j} + b_j\,\dot{q}_{i,j}
  + m_j\,g\,\ell_j\,\sin(q_{i,j}), \\
  &\qquad j=1,\ldots,6.
\end{aligned}
\label{eq:torque-proxy}
\end{equation}
where $\dot{q}_{i,j}$ and $\ddot{q}_{i,j}$ are the estimated velocity and acceleration of joint $j$ at frame $i$, and $I_j$, $b_j$, $m_j$, and $\ell_j$ are nominal coefficients. Equation~\eqref{eq:torque-proxy} produces a display-oriented torque proxy expressed in nominal $\mathrm{N\,m}$. It is based on a simplified, decoupled per-joint model with generic parameters rather than robot-specific identified dynamics. The values are intended only for qualitative visualization and must not be interpreted as measured or validated actuator torques. The proxy does not affect the commanded joint trajectory. Published joint angles are expressed in degrees and joint velocities in degrees per second, with these units declared in the trajectory metadata.

The completed record, frame sequence, derived channels, and metadata (robot model, optional gripper label, source file, unit declarations) are published to user-scoped cloud storage. The MR application synchronizes against this repository, retrieves newly published trajectories, and treats the JSON file as the authoritative motion description.

\subsection{Trajectory Fidelity and Motion Playback}

Once a trajectory has been published by the web conversion service, the MR runtime treats it as the authoritative description of the robot's intended motion. Units and conventions are explicitly stated alongside the data, ensuring that the runtime interprets the values consistently. A central design requirement is that the MR visualization faithfully reproduces the trajectory's joint configurations and playback sequence. It is intended as a kinematic visualization rather than as a validated dynamic proxy for physical hardware. Every source frame is therefore reproduced according to the declared frame rate, while interpolation provides continuous motion between frames.

The lifecycle of trajectory execution is governed by seven discrete states. From \textbf{Idle}, loading a script advances the robot to \textbf{Ready}; pressing play either enters \textbf{Adjusting} to move the robot to the start pose, or transitions directly to \textbf{Playing}. During playback, the student may pause (\textbf{Paused}), from which execution resumes exactly where it stopped. A collision triggers \textbf{Collision Halt}, freezing the trajectory until the robot is repositioned clear of the obstruction. Upon reaching the final frame, the robot enters \textbf{Complete} and returns to \textbf{Ready}. Because all transitions are driven by explicit events, the behavior remains predictable and inspectable.

Trajectory data is sampled at a fixed rate, whereas the headset renders at a higher, slightly variable rate. To bridge this gap, the runtime interpolates between consecutive trajectory frames so the robot's pose is always defined when a new image is drawn, producing smooth motion while preserving the underlying timing. The discrete gripper state is not interpolated and changes only at the explicitly programmed frames.

Before each playback, the system gently transitions the robot to the starting pose of the new trajectory. This preparatory motion prevents visual teleportation and gives the student a moment to anticipate the upcoming motion. Throughout execution, the visual pose and joint telemetry are obtained from the published trajectory record. The velocity channel contains numerically estimated and smoothed values, while the torque channel contains nominal visualization proxies.

\subsection{Kinematic--Physics Coupling}

The virtual robot operates within a world containing physical objects, simulated with a real-time physics engine so that contact, weight, and motion behave plausibly. However, the robot itself is driven by a trajectory computed without dynamics: MATLAB describes joint positions, not the required forces. Reconciling these two paradigms is the central engineering challenge of the runtime.

The system resolves this by establishing a strict boundary between the two modes. During playback, the robot is controlled entirely through joint positions rather than forces, guaranteeing it moves through the exact configurations authored by the student without drift or physics-induced jitter. Meanwhile, the physics engine independently governs everything around the robot, determining how grasped objects swing, when collisions occur, and how items fall upon release.

The two modes intersect at well-defined events. When the gripper closes around an object, the robot continues its prescribed trajectory while the physics engine attaches the grasped object so it is carried along. When a collision occurs, trajectory playback halts immediately, allowing the physics engine to take over the object's response. Consequently, the robot performs exactly as programmed, but within a physically reactive environment.

\subsection{Collision detection and object manipulation}

The simulation environment supports interaction with both detected physical surroundings and spawned virtual objects. Physical surfaces and objects detected by the Meta Quest~3 are represented by colliders in the mixed-reality scene, allowing the robot to detect contact with the user's surroundings. Spawned virtual objects additionally possess mass, colliders, and rigid-body dynamics and can be manipulated using the robot's end-effector. 

Collision detection continuously monitors for geometric intersections between the robot's links and scene objects. Upon detection, trajectory playback halts immediately, a spatial impact marker is rendered at the precise contact point, and the control panel enters a locked warning state. The student must manually reposition the robot clear of the obstruction to resume playback. This resume-in-place behavior is pedagogically valuable: rather than forcing a restart, it allows students to isolate and inspect the specific trajectory segment responsible for the collision at real scale.

Object manipulation relies on an interchangeable end-effector system and physics-enabled objects instantiated on demand. An in-scene menu provides a catalogue of spawnable objects. To keep the physics workload predictable on standalone headsets, the number of concurrently active objects is bounded, with the oldest instances removed automatically. Spawned objects can be manually picked up and arranged using standard controller or hand interactions, allowing students to set up pick-and-place scenes effortlessly. 

Grasping is executed by the gripper's own articulated joints following the states recorded in the converted trajectory. The fingers use a force-driven motion that halts upon meeting an obstruction. Contact is detected as a stall condition; when the fingers stop making progress toward their target, they hold their configuration and attach the object for transport. Upon the trajectory commanding the gripper to open, the object detaches and returns to unconstrained rigid-body simulation. 

Attaching the held object rather than retaining it through friction is a necessary consequence of control separation. A friction-based grasp would require reconciling contact forces with the arm's position-controlled joints, potentially perturbing the trajectory fidelity. The boundary between these regimes remains open to the user, allowing students to manually add or remove objects from the gripper mid-trajectory to probe how the programmed motion responds to unexpected workpiece placements.

\subsection{Interaction Model and User Experience}

The placement model is designed to eliminate initial configuration overhead. Upon instantiation, the robot automatically spawns on the nearest horizontal surface in front of the student, utilizing surface detection to avoid intersecting existing objects. This automatic placement allows the student to proceed without manually configuring the robot's initial transform.

Following instantiation, the robot can be freely repositioned using either controllers or hand tracking. With a controller, holding the grip button allows intuitive translation and rotation via the joysticks. With hand tracking, the student simply pinches and drags the robot, leveraging natural physical affordances. Both input modalities are seamlessly interchangeable.

Each robot instance is accompanied by a persistent, world-space control panel. It provides model-filtered trajectory selection, play/pause controls, and a timeline slider for non-linear motion scrubbing. A real-time joint telemetry display—presenting angles, velocities, and torque proxies—is hidden by default but accessible via a manual toggle. This is intended to reduce extraneous interface-related load during spatial observation while supporting detailed quantitative inspection on demand.

For pick-and-place tasks, a holographic cube previews the robot's intended end-effector target position prior to playback. This allows the student to visually verify the relationship between the programmed grasp location and the instantiated physical object, repositioning them as needed. Finally, the platform supports multiple simultaneously active robot instances, each maintaining independent control panels and playback states.

\subsection{Coordinated Multi-Robot Execution}

Because industrial workcells frequently utilize multiple manipulators, students must reason about shared volumes, temporal overlap, and object hand-overs. Independent per-robot playback cannot express these relationships. To address this, the runtime introduces the \textit{robot system}, a user-defined group of placed robots operated as a single synchronized unit.

A system is assembled through direct manipulation. The student enters creation mode and touches each desired robot directly with a fingertip or controller. Selected robots receive a floating numeric badge indicating their selection order. This design avoids abstract dropdown menus, keeping the grouping operation entirely spatial. 

A robot belongs to at most one system at a time. Upon joining, its individual control panel is hidden, and it is driven exclusively by the system's unified controls to prevent conflicting commands. Deleting any member dissolves the system, returning the remaining robots to independent operation.

The system offers two coordination modes. In \textit{parallel mode}, all members begin simultaneously, making temporal overlaps and potential spatial collisions directly observable. In \textit{sequential mode}, members execute one after another in their selection order, specifically supporting hand-over scenarios where one robot places an object for another to retrieve.

Before either mode executes, the system enters a staging phase. Every member simultaneously moves to the first frame of its respective trajectory, and execution only begins once every robot has arrived. This establishes a well-defined common origin for the cell. A single timeline slider represents the entire system state, allowing students to scrub synchronously backward and forward to repeatedly isolate specific hand-overs or near-collisions.

\subsection{Minimal Setup and Accessibility}

The student-facing workflow is intentionally minimal, reducing the path from a completed MATLAB program to an immersive MR demonstration into two simple steps: uploading via a browser and selecting the model in the headset. The platform automatically handles file parsing, variable detection, unit conversion, cloud storage, and floor alignment. This deliberate removal of incidental complexity ensures that students focus their attention on robotic behavior rather than on tooling.

On the web side, the conversion runs immediately in the browser and requires no MATLAB installation. MATLAB is required to author and export the original trajectory, but not on the device used to upload and convert the resulting workspace file. The platform provides plain-language diagnostics if an upload fails, enabling students to correct errors without instructor intervention. 

On the headset side, selecting an available robot model automatically places and aligns it to the floor. Once placed, the script list filters to show only compatible trajectories, preventing kinematic mismatch errors between the authored program and the virtual machine.

This cloud-based accessibility is highly advantageous for shared departmental environments. Students can author motion on a personal laptop and play it back on a shared headset without manually transferring files, while instructors can easily publish reference trajectories. 

The efficient authoring-iteration loop allows a student to correct a program in MATLAB, re-upload it, and immediately test the updated version in the headset without restarting the MR session. Ultimately, this minimalism aligns with the cognitive-load considerations reviewed in Section~\ref{RW}: by absorbing configuration overhead, the platform is intended to leave more attention available for spatial reasoning.


\subsection{Design Rationale}
The principal design decisions of the MR-Robotics LAB are grounded in the theoretical framework established in Section 2.3. The choice to render the virtual robot in the student's actual physical environment — rather than in a virtual room — reflects the embodied cognition argument that spatial reasoning is best supported when the representation shares the scale and context of the real phenomenon \cite{wilson2002six}. The decision to minimize configuration steps and reduce the workflow to upload → select → play is motivated by the finding that extraneous cognitive load introduced by interface complexity can negate the benefits of immersive media \cite{parong2018learning}. The real-time joint telemetry display is included as a deliberate pedagogical scaffold, providing the student with a data layer that connects the abstract output of the MATLAB simulation to the visible physical behavior of the robot in space — supporting the transition from computational to spatial understanding.

\section{Evaluation method}\label{SD}

The MR-Robotics LAB platform was evaluated in a single-group formative evaluation designed to characterize usability and perceived educational value, and to inform the next design iteration. 
Each participant completed a single session comprising two phases: a MATLAB desktop viewing phase followed by an MR phase, and then answered one post-experience questionnaire covering the session as a whole. 
Because participants rated the experience once rather than rating each phase on a common scale, the design yields no paired per-condition ratings and supports no comparative test between the two phases. 
It is accordingly reported as a formative, single-group evaluation rather than as a controlled comparison. 

\subsection{Participants}

Eighteen students ($n=18$) were recruited by open announcement; participation was voluntary, unpaid, and not linked to any course assessment. 
Because recruitment was self-selected, the sample is likely biased towards students already interested in robotics and immersive technology, and the reported attitudinal measures should be read with that in mind.

The sample comprised 13 undergraduate and 5 graduate students (9 female, 9 male), enrolled in the B.Sc. Information Engineering, B.Sc. Management and Data Science, and M.Sc. Management and Data Science programs. 

Prior experience with immersive technology was unevenly distributed across the sample. Five of the 18 participants (27.8\%) reported having used a VR or MR headset before, of whom 3 described their use as regular and 2 as a one-off trial.

This distribution is relevant to interpreting the usability findings in two directions. Participants without prior headset experience had to learn the interaction model during the session, so their ratings of workflow effort and confidence reflect a first-exposure learning cost that experienced users did not incur. Conversely, first-time users may also be particularly susceptible to a novelty effect, in which the immersive medium itself, rather than its spatial affordances, contributes to a positive rating.

The study was conducted in accordance with the institutional guidelines of TUM. All participants were informed in writing about the purpose of the study, the data collected, and their right to withdraw at any point without consequence, and gave informed consent before beginning the session. 
No personally identifying data were recorded; questionnaire responses were submitted anonymously and cannot be linked to individual participants.

\subsection{Procedure}

Each participant worked with a single robot trajectory that was authored in advance by the research team and provided to them as a \texttt{.mat} file. 
Each participant first observed this trajectory using the conventional MATLAB desktop visualization. 
The participant then performed the upload and conversion steps themselves on the web platform shown in Figure~\ref{fig:system-overview}, selecting the robot model and the gripper. Following automatic conversion, the trajectory became available within the MR application running on a Meta Quest~3 headset.

Participants subsequently instantiated the corresponding robot model within their physical environment and replayed the identical trajectory in MR. 
During the session, participants moved freely around the virtual robot, inspected the executed trajectory from multiple viewpoints, observed collision behavior, interacted with virtual objects, and explored the available interface features, including trajectory playback and the real-time telemetry panel.

Representative views of the evaluation session are shown in Fig.~\ref{fig:study-procedure}(a), where participants interacted with the MR-Robotics LAB using the Meta Quest~3 headset and controllers. Fig.~\ref{fig:study-procedure}(b) shows the resulting visualization, in which the virtual robot is displayed at real-world scale within the participant's physical environment.

\begin{center}
  \includegraphics[width=0.8\linewidth,keepaspectratio]{figures/image_user.jpg}
  \smallskip

  {\small (a) Participant using the Meta Quest~3 headset and controllers.}

  \medskip

  \includegraphics[width=0.8\linewidth,keepaspectratio]{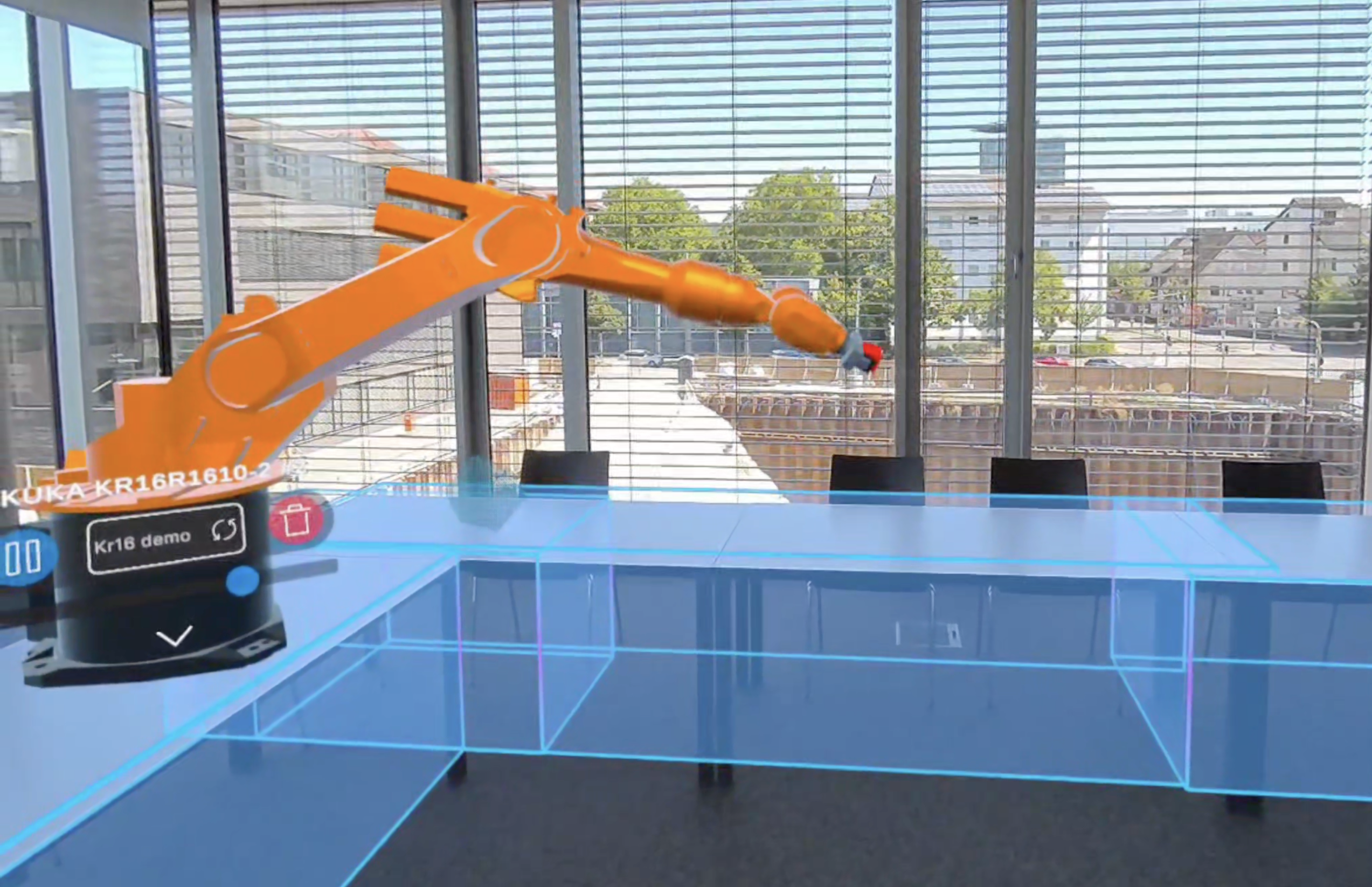}
  \smallskip

  {\small (b) Visualization of the robot within the physical environment.}

  \captionof{figure}{Representative views from the MR-Robotics LAB user
  study.}
  \label{fig:study-procedure}
\end{center}

RQ1 assesses the path from an \emph{existing} MATLAB workspace file to MR playback; participants did not author the trajectory, so the platform-independent programming stage falls outside the evaluation.

\subsection{Measures}

Participants experienced both visualization approaches using the same robot trajectory, so their questionnaire responses were collected after exposure to the screen-based simulation and the scale-correct MR. 
The questionnaire was administered once, at the end of the session, and therefore records an overall post-experience judgment rather than separate ratings of the two phases.

The anonymous questionnaire consisted primarily of seven five-point Likert-scale statements (1 = Strongly Disagree, 5 = Strongly Agree). 
The results together with the corresponding descriptive and inferential statistics are summarized in Table~\ref{tab:questionnaire}.

\begin{table*}[htbp]
\centering
\caption{Post-study questionnaire results ($N=18$). ``Distribution'' gives the count of responses at each scale point (5/4/3/2/1). $n_{\mathrm{eff}}$ is the number of non-tied responses entering each test; responses at the midpoint (3) contribute no signed rank. Exact two-sided Wilcoxon signed-rank tests against the midpoint of 3, Holm-corrected across the seven items. CI: 95\% BCa bootstrap confidence interval for the mean (100{,}000 resamples). $r_{\mathrm{rb}}$: matched-pairs rank-biserial correlation.}
\label{tab:questionnaire}

\resizebox{\textwidth}{!}{%
\begin{tabular}{p{6.6cm}cccccccc}
\toprule
Question & Distribution & Mean & 95\% CI & SD & Mdn & $n_{\mathrm{eff}}$ & $p_{\mathrm{Holm}}$ & $r_{\mathrm{rb}}$ \\
& \scriptsize 5/4/3/2/1 & & & & & & & \\
\midrule

Workflow required little effort
& 13/4/1/0/0 & 4.67 & [4.33,\,4.89] & 0.59 & 5 & 17 & $1.1\times10^{-4}$ & 1.00 \\

I could focus on understanding the robot rather than learning the software
& 9/7/2/0/0 & 4.39 & [4.06,\,4.67] & 0.70 & 4.5 & 16 & $1.5\times10^{-4}$ & 1.00 \\

I prefer verifying trajectories in MR instead of screen-based simulation
& 8/6/4/0/0 & 4.22 & [3.83,\,4.56] & 0.81 & 4 & 14 & $2.4\times10^{-4}$ & 1.00 \\

Walking around the robot improved my understanding of its workspace
& 11/6/1/0/0 & 4.56 & [4.22,\,4.78] & 0.62 & 5 & 17 & $1.1\times10^{-4}$ & 1.00 \\

I felt confident using the application without assistance
& 5/10/3/0/0 & 4.11 & [3.78,\,4.39] & 0.68 & 4 & 15 & $1.8\times10^{-4}$ & 1.00 \\

The amount of information shown was manageable
& 12/4/2/0/0 & 4.56 & [4.17,\,4.83] & 0.70 & 5 & 16 & $1.5\times10^{-4}$ & 1.00 \\

The MR visualization helped connect MATLAB values to robot behavior
& 10/3/4/1/0 & 4.22 & [3.72,\,4.61] & 1.00 & 5 & 14 & $6.1\times10^{-4}$ & 0.95 \\

\bottomrule
\end{tabular}%
}

\vspace{2pt}

\parbox{0.98\textwidth}{%
\footnotesize
\textit{Note:} $r_{\mathrm{rb}}=1.00$ indicates that every non-tied response fell above the midpoint, not that the effect is of maximal magnitude in any absolute sense. Given the pronounced ceiling in six of the seven items, these effect sizes should be interpreted as evidence of unanimous direction rather than of effect magnitude.
}

\end{table*}

The questionnaire evaluated perceived workflow simplicity, usability, confidence in operating the application, perceived information manageability, spatial understanding of robot motion, preference for MR over conventional visualization, and the relationship between MATLAB trajectory values and the observed robot behavior.

The questionnaire also collected participants' prior experience with VR or MR, their willingness to use the proposed platform in an introductory robotics course, and open-ended feedback on the platform's strengths and potential future improvements.

The instrument records only \emph{perceived} outcomes; without knowledge tests, task-performance measures, or timing data, it supports no claim about learning gain or task efficiency. 

\subsection{Analysis}

The analysis is descriptive in intent. Three features of the data collection bound what any test of these responses can establish: participants were self-selected volunteers, the session was administered by the platform's developers, and all seven items were positively worded. The full response distributions and the interval estimates, not the $p$-values, carry the substantive content of this section.

Each of the seven Likert items was tested against the neutral midpoint using an exact two-sided Wilcoxon signed-rank test. Two-sided tests were used because the direction of the hypotheses was not preregistered, despite the positively worded items implying an anticipated direction. Exact null distributions were computed from the possible sign assignments to the non-zero ranked differences, as the normal approximation is unreliable at this sample size. Midpoint responses contribute no signed rank and were excluded, so the effective sample size $n_{\mathrm{eff}}$ varies by item and is reported in Table~\ref{tab:questionnaire}. Holm's correction was applied across the seven items, and reported $p$-values are Holm-adjusted. Effect sizes are matched-pairs rank-biserial correlations ($r_{\mathrm{rb}}$), computed from the signed-rank sums using mid-ranks for tied absolute differences.

Because responses cluster near the upper bound of the scale, the distributions are markedly left-skewed. Table~\ref{tab:questionnaire} therefore reports a 95\% BCa bootstrap confidence interval for the mean (100{,}000 resamples), which adjusts for bias and asymmetry, alongside the full response distribution. Intervals were computed from the reported item response distributions with SciPy's BCa implementation, using a NumPy generator seeded at 42 for each item. Intervals whose upper bound approaches 5 indicate a ceiling rather than a precise estimate.

These tests indicate that the response distributions differed from the neutral midpoint under the assumptions of the signed-rank test. They do not speak to the magnitude of any effect, and because the design produced no paired per-phase ratings, they cannot show that MR outperformed the MATLAB desktop phase.

Free-text responses were analyzed using an inductive thematic coding approach. Two researchers independently open-coded all responses without a predefined codebook, then reconciled discrepancies through discussion and grouped the agreed codes into four themes: workflow usability, spatial understanding, educational value, and requested extensions. Given the small corpus of short questionnaire responses, this analysis serves as an interpretive complement to the quantitative results rather than as independent evidence. 

\section{Findings}\label{F}

The findings are organized according to the three research questions introduced in Section~\ref{I} and are reported as means, agreement percentages, and response distributions.

\subsection{Workflow simplicity and usability (RQ1)}

RQ1 asked whether students can move from a completed MATLAB trajectory to MR playback with low setup effort. All four usability items were rated positively. Perceived effort received the highest rating of the instrument ($M=4.67$, 94.4\% agreement), followed by manageable information load ($M=4.56$, 88.9\%), the ability to focus on the robot rather than the software ($M=4.39$, 88.9\%), and confidence in operating the application unassisted ($M=4.11$, 83.3\%). The information-load item records perceived manageability and is not a validated cognitive-load measure.

Free-text responses were consistent with these ratings. Participants described the transition from upload to MR playback as seamless and valued the ability to place and inspect a robot without configuration. Their usability suggestions (an introductory tutorial, clearer interaction prompts, and easier scrolling through model and script lists) focused on interface refinements rather than obstacles to completing the workflow. Taken together, these results provide formative evidence that participants perceived the workflow as simple and manageable, thereby supporting RQ1 at the level of perceived usability.

\subsection{Spatial understanding (RQ2)}

RQ2 examined which spatial properties of robot motion became accessible under scale-correct MR playback. The strongest result was for moving around the robot to inspect its motion from different viewpoints ($M=4.56$, 94.4\% agreement). Participants also reported that MR helped them connect numerical MATLAB values to visible robot behavior ($M=4.22$, 72.2\%); this was the only item to receive a response below the midpoint and showed the widest dispersion of the seven ($SD=1.00$). Preference for verifying trajectories in MR over screen-based simulation was likewise positive ($M=4.22$, 77.8\%), with four neutral ratings and no disagreement. Because the two conditions were not rated separately on a common scale, this item records a stated preference rather than a paired comparison.

Open-ended responses identified the robot's true scale, its relationship to the surrounding room, and multi-perspective observation as the most useful aspects of the experience, with specific mention of reach, workspace boundaries, trajectory accuracy, and collisions. One participant noted that moving around the robot made it possible to ``verify if there are any issues with the trajectory''; another observed that overlaying motion on the real environment made collisions easier to spot. Comments on collision boxes, real-room appearance, and the placement of multiple robots suggest that the physical context provided information that participants did not find equally accessible on-screen.
These findings provide formative support for RQ2 regarding participants' perceived spatial understanding.

\subsection{Interest in robotics education (RQ3)}

Asked whether they would test their own scripts on the platform in a future introductory robotics course, 15 of 18 participants (83.3\%) answered yes, 3 (16.7\%) were undecided, and none declined.

Requested extensions (additional robot models, coordinated multi-robot interaction, shared object manipulation, live parameter adjustment, variable playback speed, richer environments) indicate interest in more complex scenarios than those demonstrated in the workflow. 
Asked which complementary tools they would want in such a course, participants named physical robots, additional headsets, and the Apple Vision Pro.

These results support RQ3, though whether the interest persists beyond a single session would require a longer classroom deployment to establish.

\section{Discussion}\label{D}

\subsection{What MR adds to screen-based simulation}

The pattern across the RQ2 items and free-text responses points to a single affordance: the viewer's own position becomes the camera, and the surrounding room supplies a metric reference against which the robot is read. 
These remain reported impressions rather than a demonstrated advantage. Their consistency across the sample and their convergence with the study~\cite{sibilska2023analysis} comparison of immersive against 2D robot visualization make the direction credible, but confirming it would require rating both conditions on a common scale.

MATLAB remains the appropriate environment for trajectory generation, numerical analysis, and debugging; MR adds spatial verification between authoring and hardware execution. Participants' interest in using physical robots \emph{alongside} the platform is consistent with that reading.

\subsection{Design implications}

Three decisions appear transferable to other educational MR tools.

\emph{Leave the authoring toolchain untouched.} Students write trajectories exactly as their course already requires, and the platform absorbs the conversion. This costs no curricular time and removes the need to teach a second authoring environment alongside the domain content.

\emph{Treat scale fidelity as a correctness requirement.} The spatial reasoning described above depends on link dimensions surviving URDF conversion unaltered and on anchoring holding the model to a stable room pose; a system that renders a plausible but unscaled robot forfeits the property that distinguishes it from a desktop viewer.

\emph{Decouple preprocessing from the headset.} Browser-based conversion with cloud storage lets students author, upload, and replay across different devices, which matters where headsets are shared departmental equipment.

\subsection{Challenges and trade-offs}

Two practical issues surfaced. Perceived visual clarity varied between participants, suggesting that headset fit and individual vision affect the experience in ways a single-device evaluation cannot characterize. Participants new to MR also needed a short familiarisation period with the controller interaction before working comfortably, which points to guided onboarding as a worthwhile addition.

The requested extensions redirected development. Coordinated multi-robot interaction, the most frequent request, was implemented as the robot-system mechanism in Section~\ref{ARLab}, and the model library was extended beyond KUKA to include the UR5e, thereby confirming that the URDF import path generalizes across manufacturers.
This illustrates the formative, evaluation-driven development of the system: the evaluation not only characterized the artifact but also informed its subsequent development. Neither addition has been evaluated, and it remains open whether grouped playback aids reasoning about multi-arm workcells more than independent playback.

\section{Limitations}\label{L}
Three constraints bound the scope of these findings.

\textbf{Study design.} The evaluation involved 18 participants at a single institution in one session, with the MATLAB phase always preceding the MR phase and no control group. The design is formative: it was intended to surface usability obstacles and design requirements rather than to estimate the medium's effect. The findings, therefore, report participants' experiences rather than measured differences between conditions, and order effects, novelty, and the positively worded items are inseparable from that experience, despite the use of anonymous responses. A multi-institution study with counterbalanced conditions, validated instruments (SUS or UEQ for usability, NASA-TLX for load), and an objective performance measure would settle the comparative question this study leaves open.

\textbf{Trajectory conversion.} The converter reads MATLAB workspace (\texttt{.mat} file) variables rather than executing MATLAB, which removes any license or runtime dependency on the server and keeps the platform lightweight. In exchange, trajectories must be saved beforehand as a finite two-dimensional matrix with six joint columns and an optional seventh gripper column, in that order, and variables held in unconventional structures must be restructured before upload. Arbitrary kinematics, configurable joint mappings, and additional robotics ecosystems are a direct path for future work.

\textbf{Mixed reality application.} The runtime inherits the same format assumptions, so mobile bases and dual-arm systems would require work on both the conversion and visualization sides. Scene state is not yet persisted between sessions, and the interaction design was developed and evaluated on a single headset (Meta Quest~3) using its native hand tracking and controllers, leaving portability to other MR platforms to be confirmed. Persistence and multi-user support would extend the platform from a single-session demonstration towards sustained classroom use.

\section{Conclusion}\label{C}

This paper presented an end-to-end workflow that bridges MATLAB-based robotics simulations and MR visualization through a standardized JSON trajectory format and a web-based conversion platform. By requiring only an exported MATLAB workspace, the proposed system enables students to convert simulated robot trajectories into immersive MR experiences without modifying their existing simulation workflow. This lowers the barrier to adopting MR as a complementary educational tool in robotics courses.

The formative evaluation with 18 engineering students indicated that the proposed workflow was well received by this sample. Participants reported that the web application required little effort to use and that the MR visualization helped them understand robot motion and workspace relationships. Furthermore, a large majority expressed interest in using the platform in future robotics courses. These findings concern perceived usability, understanding, and willingness to adopt the platform.

The proposed architecture also provides a flexible and reusable interface between robotics simulation software and immersive visualization environments. By introducing a standardized JSON representation of robot trajectories, the workflow decouples trajectory generation from visualization, enabling the incorporation of additional robot models, simulation environments, and visualization platforms with relatively minor modifications.

Future work will focus on expanding support for robots with arbitrary numbers of joints, improving the flexibility of trajectory conversion, integrating additional robotics software ecosystems, and enabling real-time communication between simulation environments and MR applications. Larger-scale and longitudinal evaluations will also be conducted to investigate the long-term educational impact of MR on robotics learning.

Overall, the presented workflow demonstrates the feasibility of connecting established robotics simulation tools with room-scale MR through a lightweight web-based pipeline. The results suggest that students perceive value in inspecting robot motion at real scale while making only minimal changes to their existing workflow. Whether the platform improves learning or task performance relative to conventional visualization remains to be established through a controlled evaluation.

\section*{Acknowledgments}

This research was financially supported by the TUM Campus Heilbronn \textit{Incentive Fund 2025} of the Technical University of Munich, TUM Campus Heilbronn. We gratefully acknowledge their support, which provided the essential resources and opportunities to conduct this study.

\pagebreak
\printcredits

\bibliographystyle{cas-model2-names}

\bibliography{00Paper}

\end{multicols}
\end{document}